# One Patient, Many Contexts: Scaling Medical AI with Contextual Intelligence

Michelle M. Li[1,2,†], Ben Y. Reis[1,2,3,4], Adam Rodman[5], Tianxi Cai[1,6], Noa Dagan[2,7,8], Ran D. Balicer[2,7,9], Joseph Loscalzo[10], Isaac S. Kohane[1,2], Marinka Zitnik[1,2,4,11.12,†]

[1] Department of Biomedical Informatics, Harvard Medical School, Boston, MA, USA
[2] The Ivan and Francesca Berkowitz Family Living Laboratory Collaboration at Harvard Medical School and Clalit Research Institute, Boston, MA, USA
[3] Predictive Medicine Group, Computational Health Informatics Program, Boston Children's Hospital, Boston, MA, USA
[4] Harvard Data Science Initiative, Cambridge, MA, USA
[5] Division of General Internal Medicine, Department of Medicine, Beth Israel Deaconess Medical Center, Boston, MA, USA
[6] Department of Biostatistics, Harvard T.H. Chan School of Public Health, Boston, MA, USA
[7] Clalit Research Institute, Innovation Division, Clalit Health Services, Ramat-Gan, Israel
[8] Software and Information Systems Engineering, Ben Gurion University, Be'er Sheva, Israel
[9] Faculty of Health Sciences, School of Public Health, Ben Gurion University of the Negev, Be'er Sheva, Israel
[10] Department of Medicine, Brigham and Women's Hospital, Harvard Medical School, Boston, MA, USA
[11] Kempner Institute for the Study of Natural and Artificial Intelligence at Harvard University, Allston, MA, USA
[12] Broad Institute of MIT and Harvard, Cambridge, MA, USA
[†] Correspondence: michelleli@g.harvard.edu, marinka@hms.harvard.edu

## Abstract

Medical AI, including clinical language models, vision-language models, and multimodal health record models, already summarizes notes, answers questions, and supports decisions. Their adaptation to new populations, specialties, or care settings often relies on fine-tuning, prompting, or retrieval from external knowledge bases. These strategies can scale poorly and risk contextual errors: outputs that appear plausible but miss critical patient or situational information. We envision context switching as a solution. Context switching adjusts model reasoning at inference without retraining. Generative models can tailor outputs to patient biology, care setting, or disease. Multimodal models can reason on notes, laboratory results, imaging, and genomics, even when some data are missing or delayed. Agent models can coordinate tools and roles based on tasks and users. In each case, context switching enables medical AI to adapt across specialties, populations, and geographies. It requires advances in data design, model architectures, and evaluation frameworks, and establishes a foundation for medical AI that scales to infinitely many contexts while remaining reliable and suited to real-world care.

## Introduction

Medical foundation models now match or surpass clinicians on benchmark tasks, such as summarizing notes, answering board-style questions, and assisting in diagnosis[1–4]. They streamline workflows[5,6] and report accuracy that rivals expert judgment in controlled settings[7–9]. These successes raise expectations for clinical use[10] and suggest that models can perform reliably even without reasoning in the same way as clinicians. Yet strong benchmark performance can collapse in real-world care[11].

Patients present with comorbidities, rare diseases, and evolving treatment standards that rarely appear in training data[12,13]. In these cases, model outputs often sound correct but miss critical contextual details[14,15]. These contextual errors expose a central limitation: models recognize statistical patterns[16,17] but struggle to adjust their reasoning when the clinical context shifts.

This gap stems from a two-stage paradigm of large-scale pretraining followed by narrow, task-specific fine-tuning on datasets with curated labels for tasks[1,2,7], such as triage, diagnosis, or prognosis[1,2,7,18]. Performance depends on how well the training distribution matches the deployment setting. When distributions vary across hospitals, populations, or specialties, performance drops[17,19,20]. Prompting[2,21], retrieval, or multimodal inputs[5,8] can improve results, but these strategies require repeated fine-tuning, which cannot scale across all possible clinical contexts.

As AI moves into healthcare, context-awareness becomes urgent. Models must adapt automatically to differences in users, health system, geographies, diseases, and populations[22]. Without such adaptation, models rely on manual adjustment, which limits scalability and introduces errors[14,15]. Precision medicine demands care plans tailored not only to the patient but also to the constraints of the health system that delivers the care[22,23].

We envision context-switching as the defining paradigm of medical AI. It refers to a model that adjusts reasoning and outputs as the clinical context shifts across specialty, population, task, data availability, or local practice (**Figure 1**). Context-switching does not require new knowledge; it applies existing knowledge flexibly during inference (**Box 1**). A model that interprets chest X-rays in a tertiary hospital should still function in a clinic with partial data. A model trained on adult cardiology should condition on physiological differences to extend to pediatric cases. A documentation model should shift its tone when writing for patients versus clinicians. These adjustments occur during inference, not through retraining. Rather than operating as rigid pipelines, medical AI must function as adaptable models that respond as variation arises.

Context-switching builds on three pillars. Data strategies must capture context-specific signals from patient records and medical knowledge. Models must adjust their reasoning at test time to differences in data, user roles, and disease settings. Evaluation must expose how models behave across real-world variability and identify failure modes that guide more generalizable design. These directions chart a path for medical AI that learns to reason in context rather than remaining tied to static distributions. We envision context-switching as the guiding paradigm for the future of medical AI, one that aligns model behavior with the complexity of clinical practice and the diversity of patients.

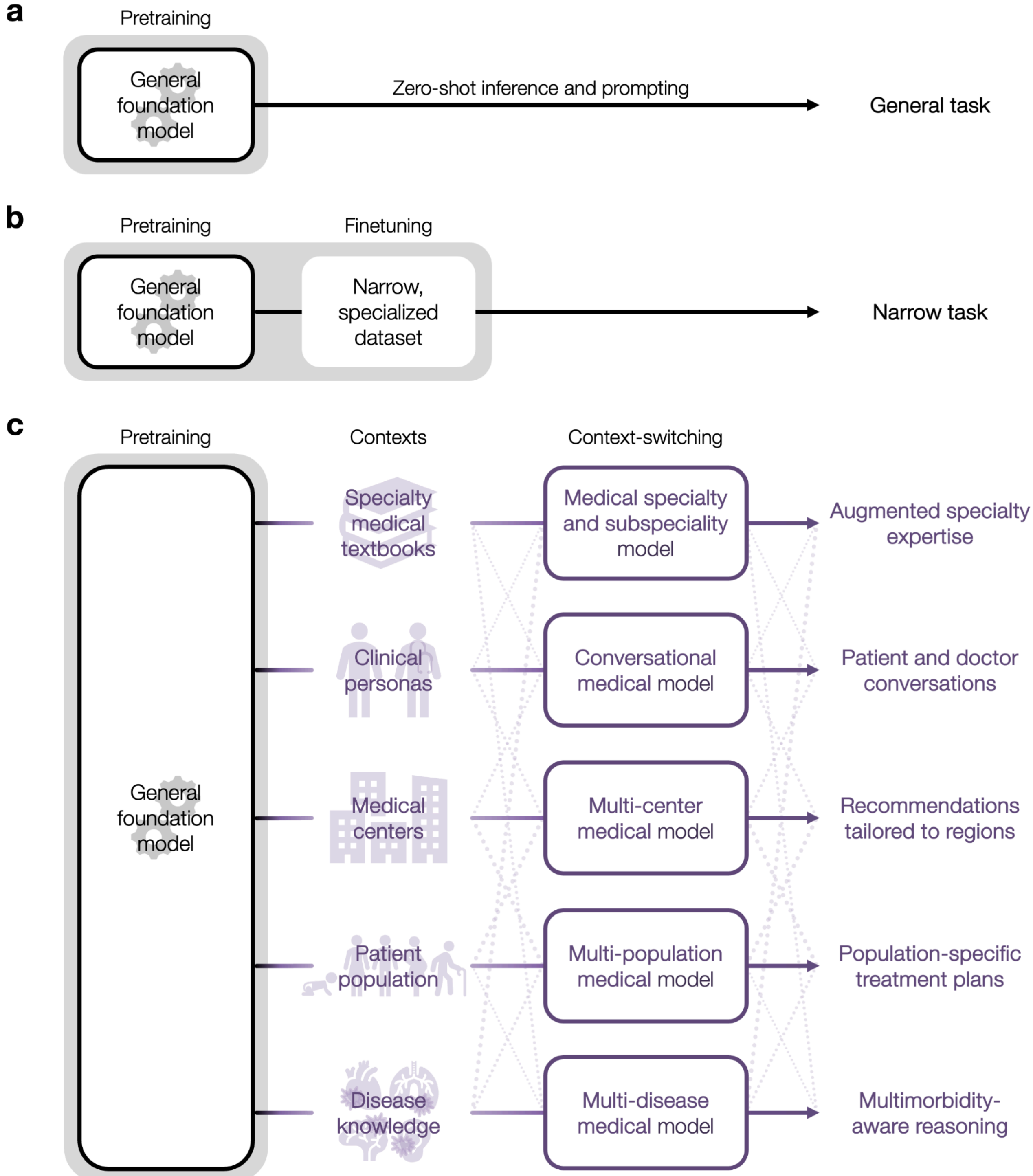

**Figure 1. From pretraining to contextual adaptation in medical AI. (a) Pretrained models in the general domain.** Foundation models trained on large, general-domain data can perform a wide range of tasks but often fail to generalize in specialized medical settings. Their performance declines when applied to domains that require localized or domain-specific knowledge. **(b) Fine-tuning, prompting, and retrieval from external knowledge bases.** Fine-tuning aligns general models with specific medical tasks. However, performance gains

diminish when distribution shifts are large, such as those due to different diseases, populations, or care settings, making fine-tuning inefficient and hard to scale. **(c) Context-switching in medical AI.** Context-switching models adjust in real-time to changes in specialty, disease, user role, or data availability without retraining. This shift supports broader clinical utility by reducing reliance on labeled data and serving many tasks, users, and clinical settings. Context-switching models adapt to infinitely many medical specialties, healthcare roles, diseases, and populations.

**Box 1: Early strategies for context-switching in medical AI.**

Several approaches illustrate how models begin to adapt their behavior across contexts:

**Prompt engineering.** Input instructions supply the missing context and narrow the search space of possible outputs.

**In-context learning (ICL).** Zero- or few-shot examples in the prompt show the model how to perform a task in a new setting.

**Context discovery.** Algorithms identify, retrieve, and compress the most relevant evidence from external data sources, rerank it by user constraints, and feed it into the model.

**Supervised fine-tuning (SFT).** Models update on curated datasets to perform a target task. Parameter-efficient fine-tuning achieves similar performance with fewer resources; full fine-tuning is reserved for cases that require large behavioral shifts.

**Post-pretraining adaptation.** Multi-round refinement adjusts model behavior and aligns outputs with user intent. Techniques include instruction tuning to follow instructions, domain- and task-adaptive pretraining to capture specialized vocabulary and patterns, and adaptations that teach models to use tools and call functions. Preference optimization, often implemented through reinforcement learning from human feedback, further aligns outputs with user expectations.

**Agent models.** Autonomous large language model agents plan, reason, and act across contexts. Specialized components for planning, memory, tool use, or task decomposition collaborate on multi-step workflows and adjust outputs based on feedback from one another.

## Prompting and fine-tuning models for user-defined contexts

Foundation models facilitate interactions between computational models, providers, and patients[5,6,8,9]. For example, personalized chatbots extend timely care to patients with rare, complex, or stigmatized conditions who often face delays in access[4]. These models leverage natural language as a universal interface between humans and machines[24], enabling context-switching through prompt-based interactions (**Box 1**). A prompt may include input-output examples that the model uses to infer the expected structure and content of the task[25–29]. When a sufficient number of examples are available, the model can be adapted through fine-tuning to improve performance on the new task[30–33]. Pretraining followed by prompting and fine-tuning represents an initial step toward context-adaptive models, but remains limited because adaptation depends on carefully curated examples and repeated fine-tuning for each new task[17,19,20,34] (**Boxes 1-2**). In medical contexts, aligning a model's training with the intended output is difficult because clinical presentations involve subtle patterns, such as comorbidities or atypical findings, that general prompts rarely capture[12,13,35]. Prompting also depends on human-computer interaction factors, including the clinical expertise of the user, cognitive biases[36], and the time available to craft and iterate on prompts[37]. These constraints intensify in high-pressure settings, such as emergency care[37], and in resource-limited settings, such as rural clinics. Prompting frameworks that automatically optimize prompts without updating model parameters reduce manual effort[38–40] and improve performance on medical question-answering tasks[41–44], indicating that context-switching can emerge from optimized prompting and in-context learning. However, even when models perform well in isolation, they may not work effectively when used alongside clinicians[11,45]. Large language models often encode cognitive biases and display sycophantic behavior[46,47], distorting responses and unpredictably influencing the clinicians who prompt them.

Prompting a foundation model while relying solely on its parametric knowledge limits its capacity to handle distribution shifts in target labels[5,6,8,9,34], such as updates to disease classifications, evolving diagnostic criteria, or newly recognized syndromes[12,13,35]. The verification of generated outputs through retrieval-augmented generation (RAG) and knowledge graphs[48] can improve contextual relevance in medical question-answering tasks[49–52], indicating their potential to enable context-switching in real-time settings. Retrieval-based methods reduce hallucinations, improve accuracy, and increase efficiency, but require context-relevant data in the vector database[53–56] (**Box 1**). Such data are scarce for patients from underrepresented groups, from low-resource regions, with rare conditions, or who speak less widely used languages[57]. Although effective, fine-tuning requires curated data and risks overfitting to narrow tasks while degrading performance on others (**Boxes 1-2**). To improve generalization, methods have incorporated large volumes of unlabeled data through unsupervised or self-supervised learning. But unlabeled data alone cannot replace targeted supervision in contexts that demand reasoning over complex relationships, disambiguation of subtle clinical signals, or alignment across diverse modalities. Supervised examples define task boundaries, highlight relevant features, and direct the model toward clinically actionable information. Without such guidance, models may learn spurious correlations or overlook rare or inconsistently recorded phenotypes. Only models that adapt at inference can scale safely across the diversity of clinical settings.

These limitations, along with patient privacy concerns, curb the scalability of prompting and fine-tuning as well as their potential to become strategies for adapting models to real-time clinical variation[58]. Sharing large language models trained on patient data is often not possible due to concerns about privacy leakage and re-identification[59]. These restrictions prevent health systems from training models on local data and transferring them across other health systems for continued fine-tuning. As a result, models cannot be incrementally trained across diverse patient populations and care settings, which has been a longstanding practice of medical AI. Without privacy-preserving mechanisms, collaborative model development across health systems remains infeasible. By dynamically tailoring outputs to different contexts (e.g., patients, health systems) at inference, context-switching can enable context-aware reasoning while preserving patient privacy.

Foundation models that rely only on prompting and fine-tuning recall medical information but rarely reason or adapt to new tasks without additional training (e.g., post-pretraining adaptation) or manual intervention (e.g., test-time scaling). Clinical practice requires models that apply learned principles to new situations by adjusting their reasoning dynamically[36,60]. Context-switching moves beyond recall to support reasoning and decision-making grounded in the patient's context[61].

**Box 2: Opportunities for context-switching in medical AI.**

Context-switching is an emerging paradigm with multiple opportunities for innovation. Five open challenges in medical AI highlight where context-switching can have immediate impact:

**Insufficient context in the query.** When context is missing, models can struggle to narrow the search space and often reach incorrect conclusions. Current models fill these gaps from parametric knowledge. Context-switching identifies missing information and prompts the user to supply additional details, producing more accurate responses.

**Static context encoding.** Retrieval methods often encode external knowledge in ways that restrict extrapolation from small fragments (e.g., sentences) across large sources (e.g., dense documents). Dynamic, context-aware encoding preserves nuance and prevents evidence dilution, enabling more reliable reasoning.

**Catastrophic forgetting.** In continual and lifelong learning, updating model parameters can improve domain-specific performance while degrading general capabilities. Analogous to continuing medical education, context-switching supports efficient cycles of learning and unlearning to refine knowledge without erasing prior competence.

**Reward hacking.** In reinforcement learning, models may exploit loopholes to maximize rewards while failing to achieve the intended outcome, producing misaligned or harmful behavior. Context-switching designs reward functions that are multi-faceted, multi-step, and sensitive to context, reducing this risk.

**Compound error.** Generative and agentic models accumulate errors across steps, leading to flawed outputs. Context-switching detects when errors exceed tolerable limits, identifies the step at which failure arises, and revises the reasoning process before continuing.

## Guiding the generation toward context-aware clinical outputs

Generative models create new content (e.g., patient trajectories[62,63], medical images[64], or free-text explanations[65]), and must adapt their outputs based on the clinical setting, user background, and the scale of biomedical information. Generative models often rely on conditional generation, where outputs are shaped by structured prompts, patient features, or other contextual signals. Transformer-based decoders, flow-matching models, diffusion models, and variational autoencoders are common backbones, with conditional representations steering generation across different domains[66–68]. Beyond prompt engineering, in-context learning, and parameter-efficient finetuning, post pretraining model adaptation strategies can guide the model's behavior and outputs toward the desired contexts (**Box 1**).

User-adaptive decoding methods allow the model to tailor outputs based on recipient characteristics, such as medical expertise or health literacy level[69–71]. This may involve context-aware language modeling with controlled generation, embedding user profiles into prompts, or adjusting decoding strategies (e.g., temperature, top-k sampling) to modulate output specificity and tone[72,73]. A model that generates patient instructions, for example, should adjust its tone, vocabulary, and level of detail depending on whether it is addressing a specialist physician, a medical trainee, or a patient with limited health literacy[69,70]. When generating summaries for a care team, the model should distinguish between electronic health record (EHR) documentation, specialist referrals, and insurer justifications. Context-switching supports these shifts by interpreting task-specific cues and adjusting the output structure accordingly. Domain- and task-adaptive pretraining (**Box 1**) can refine a model's vocabulary and patterns to reflect the intended users (e.g., physician, trainee, patient) or use cases (e.g., EHR documentation, medical referrals, insurance documentation). Although first developed in general-domain language models, context-switching strategies extend naturally to medical AI, where language serves as the interface between humans and machines[24].

## Learning how and when to context-switch across data modalities

Multimodal models are designed to process and integrate diverse data types[32,33], including structured EHRs, unstructured clinical notes, medical knowledge (e.g., textbooks, research articles, and guidelines), laboratory test results, imaging modalities (e.g., radiology and pathology)[30,31,74,75], and genomic data[76–78]. Late fusion models[79–82] encode each modality independently before integrating their representations; early fusion encodes raw inputs jointly; and hybrid approaches combine both[79,80]. Cross-modal attention mechanisms align semantically related information, while modality-specific encoders retain modality-specific information[30,32,74]. To handle missing or variable inputs, some models use gating or masking strategies[77,78], modular encoders, and shared latent spaces that add new modalities without retraining from scratch. Using language as the interfacing modality, these models switch across modalities to integrate information in different clinical contexts (**Box 1**).

Multimodal context-switching enables models to adjust their use of different data modalities dynamically at inference time based on the clinical setting and available inputs[83]. For example, a model may rely on histopathology images to classify a tumor when slides are available, but shift to lab values and clinical notes when imaging is missing. Beyond substituting one modality for another, multimodal context-switching allows models to combine data types that are not co-observed during training, such as ECG waveforms and genetic variants in cardiology risk assessment. This behavior requires models to infer which modalities are most informative, detect redundancy or missingness, and reweight their reasoning in real-time. Retrieval-based context discovery can realize this capability by considering user-defined metrics on clinical informativeness in the reranking step (**Box 1**). Preference optimization can learn a reward function that aligns model outputs with human preferences based on a set of human-annotated examples (**Box 1**). Further supporting this flexibility can depend on mechanisms like attention-based modality selection, uncertainty-aware fusion, and mixture-of-modality experts that specialize based on the task and context[84–87].

Without context-switching, multimodal models remain brittle and fail to operate across variable medical data modalities. Multimodal context-switching instead learns to reason over incomplete, asynchronous, and novel combinations of modalities by aligning their latent representations without relying on rigid predefined structures. This makes multimodal context-switching well-suited for clinical reasoning, where data are incomplete and decisions must be made with partial and evolving information. As medical data grow more complex, models that navigate uncertainty without rigid constraints are needed to generate reliable predictions from incomplete, inconsistent, or rapidly changing data.

## Reasoning in context through multi-step analysis

Hybrid models offer a modular design for achieving context-switching across tasks, modalities, and user roles. These models combine ideas from multimodal learning, where models jointly interpret diverse data types, and generative modeling, where outputs are conditioned on specific contextual signals. Compound models, such as agents, provide a modular foundation for context-switching (**Box 1**). In these models, different subagents are trained to specialize in distinct tasks, data modalities[85,86], or clinical domains[88,89], and are orchestrated to solve problems that require context-specific adaptation[90]. For example, a dermatology case involving both imaging and text may trigger a vision agent specializing in skin lesion classification and a language agent that processes clinical notes describing symptom progression. Routing decisions are either optimized during training or dynamically inferred at test time based on features of the input or task prompt. This approach enables efficient specialization and reduces unnecessary computation, but relies on robust mechanisms to manage transitions between components.

Agents are goal-directed models that decompose complex tasks into subtasks, plan actions, and interact with data, tools, or users in real-time[88–90] (**Box 1**). Agent models typically consist of multiple specialized agents, each trained to perform a distinct function (e.g., information retrieval, question answering, decision support, patient communication). Context-switching in this setting refers to selecting and orchestrating the appropriate agent or submodel based on features of the task, user role, or available data. For example, an agent coordinating cardiovascular risk assessment may activate one subagent to extract patient history, another to synthesize lab results, and a third to generate a summary tailored to a clinician or patient.

Rather than relying on single agents to generalize across tasks, a multi-agent approach applies context-switching by routing tasks to the most relevant experts. A supervisory or executive agent interprets input features, such as the type of clinical query, user background, or data modality, and assigns subtasks to specialized agents. Coordination can be implemented through gating functions, learned routing policies, or rule-based controllers that use structured metadata to direct workflows. For example, a pharmacovigilance workflow may use agents for adverse event detection, regulatory interpretation, and causal inference, with an executive agent selecting which experts to engage depending on whether the task is clinical trial monitoring or post-market surveillance. In this setting, context-switching arises not from each agent altering its reasoning strategy, but from routing and assigning the right expert agents to the clinical context.

Models that perform multi-step analysis represent a key direction for context-switching in medical AI. Reasoning models learn inference procedures that chain intermediate steps to address tasks, such as diagnosis, treatment selection, or risk prediction[91–93]. Reasoning models, often built on large language models fine-tuned with reinforcement learning or supervised traces, learn both the final output and the sequence of reasoning steps that generate it[93,94] (**Box 1**). This abstraction makes reasoning models well-suited for context-switching, since reasoning strategies can often transfer across tasks. For example, given a patient's symptoms, a reasoning model might infer a molecular diagnosis by sequentially identifying phenotype-gene associations, retrieving pathogenic variants, and ranking candidate genes. The same model could then apply a similar reasoning process to identify treatment options, retrieve drugs that target the implicated gene, evaluate drug indications, and prioritize therapies based on safety and efficacy. Context-switching in reasoning models occurs not by changing the model architecture, but by adjusting the sequence and structure of inference steps based on the input prompt. In a novel context, reasoning models can facilitate multi-turn collaboration[34] with users, in which the model iteratively identifies the gaps in its knowledge or the given user instruction and prompts the user to provide additional guidance (**Box 2**).

A central challenge in developing reasoning models is designing reward functions that guide the model through intermediate steps of clinical reasoning[94]. These functions must assess the final prediction as well as the clinical relevance and plausibility of each step. For example, in differential diagnosis from patient symptoms[89], a reward function might penalize inconsistencies between symptoms and candidate diagnoses, reward logical progression in differential diagnosis from common to rare conditions, and prioritize hypotheses that fit the patient's demographic and clinical context. This structure encourages models to revise reasoning when context shifts, such as when new symptoms appear or the task changes from screening to treatment planning[95]. Reward functions must align with clinical objectives, including reducing diagnostic error, avoiding unnecessary procedures, and improving continuity of care. Reasoning models that incorporate such context-aware rewards can adjust how they interpret and prioritize information and adapt to clinical goals and constraints.

**a** Switching across medical specialties and disease contexts

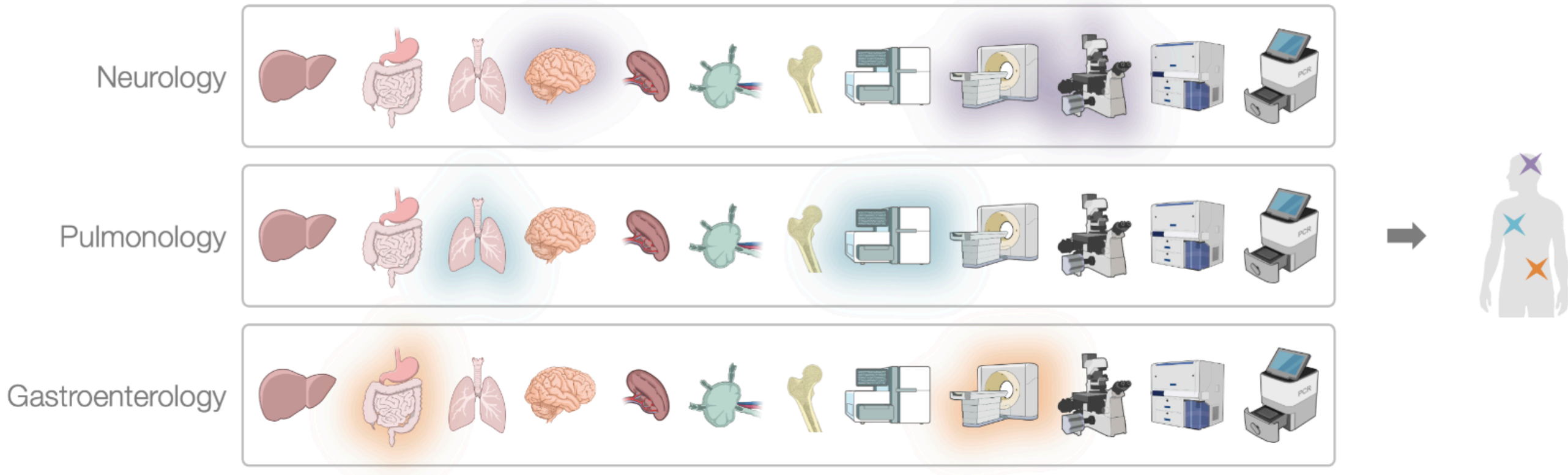

**b** Switching across geographies and patient populations

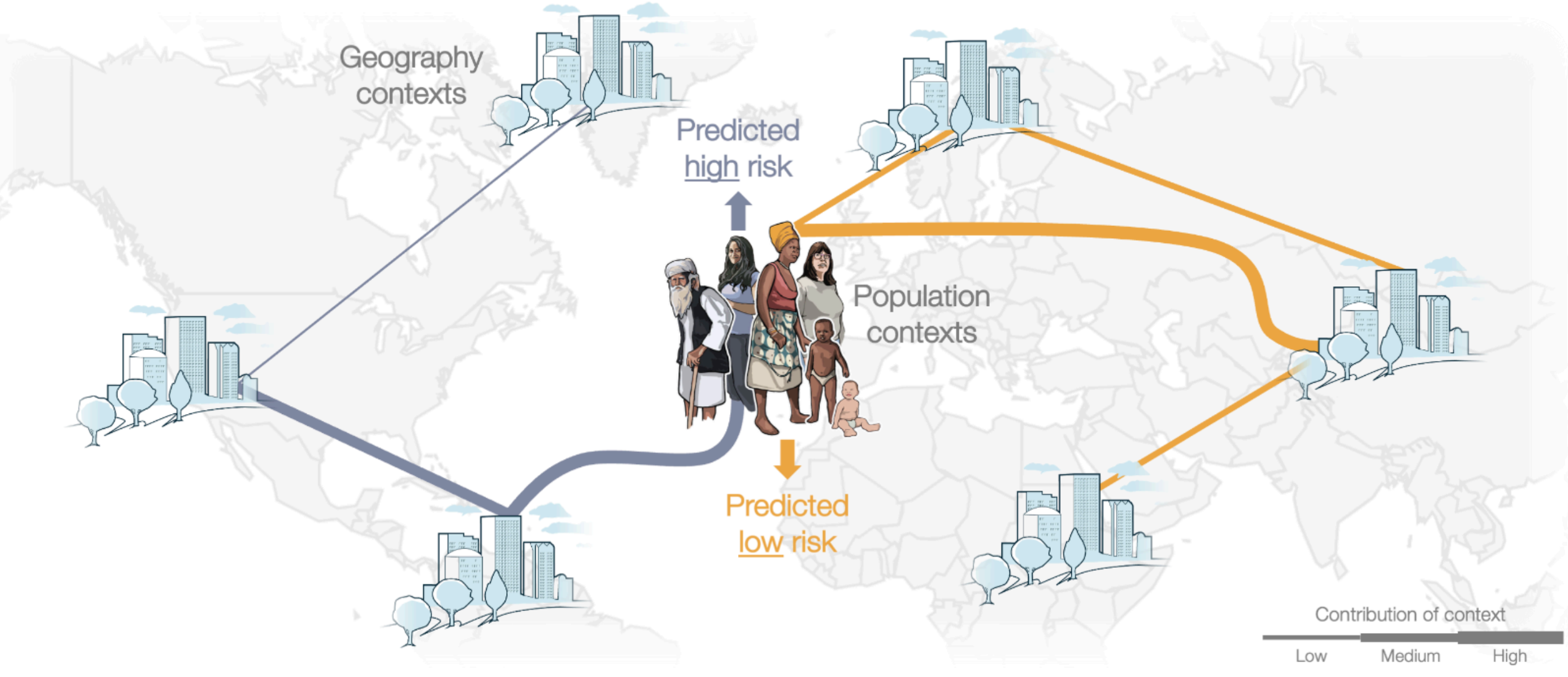

**c** Switching across healthcare roles

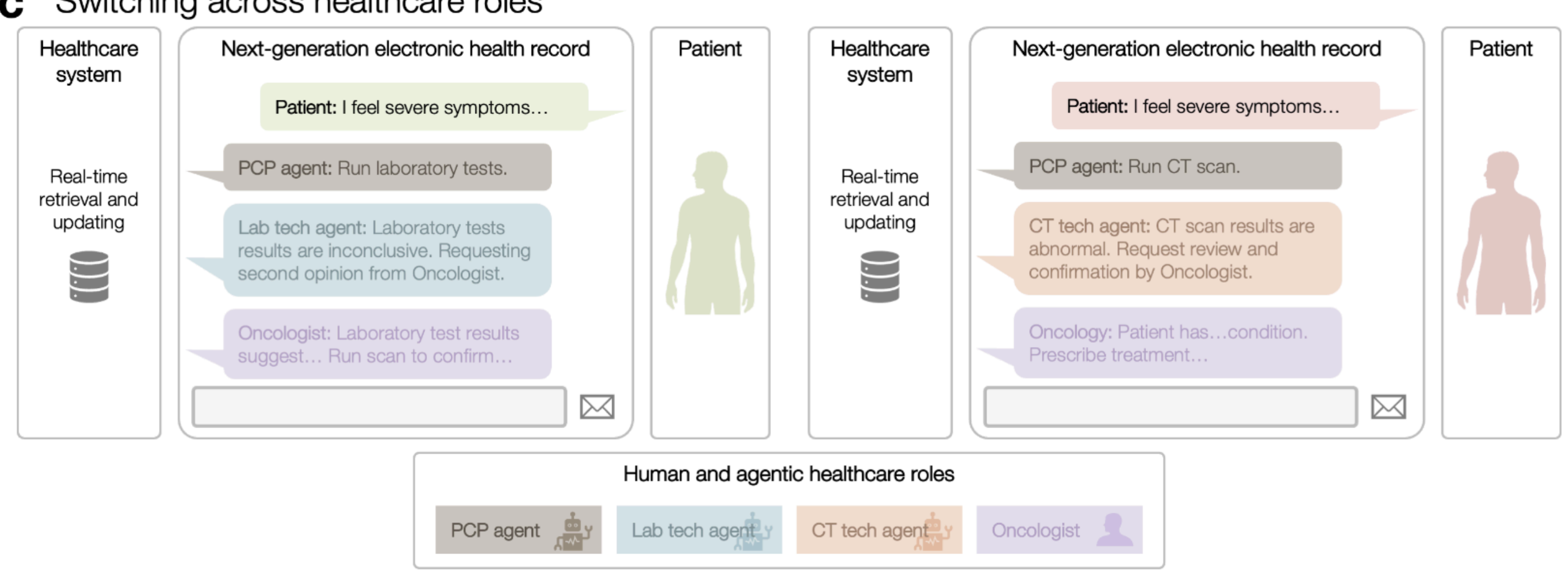

**Figure 2. Context-switching across medical specialties and diseases, geographies and populations, and healthcare roles. (a)** A context-switching model for medical specialties and diseases can be a multimodal generative model that identifies and integrates the most relevant data modalities (e.g., histopathology images, CT scans, whole genome sequencing) and clinical specialties (e.g., neurology, pulmonology) to guide diagnosis and treatment. **(b)** Adapting to geographic and population-specific needs can involve a generative reasoning model capable of multi-step inference, capturing both high-level reasoning (e.g., shared disease risk factors) and fine-grained variation (e.g., subpopulation-specific modifiers). **(c)** A generative agent system, where each agent represents a distinct healthcare role (e.g., nurse, primary care physician, specialist), can flexibly engage the appropriate expert(s) for a given case, involve the patient throughout the decision-making process, and continuously update outputs using real-time access to the patient's electronic health records. Illustrations adapted from NIAID NIH BIOART Source.

## Why medical AI must adapt to who, where, and what it serves

Medical AI must do more than process diverse inputs or scale across computing infrastructure. It must adapt to the contexts of care, aligning with patient needs, care settings, and clinical workflows. Specialty, geography, population, and access expose the limits of static models and make context-switching a requirement for medical AI.

### Bridging medical specialties and disease contexts

Patients often face fragmented care in siloed health systems or resource-limited regions, leading to diagnostic odysseys and delays or errors in care[96,97]. Context-switching models enable clinicians to draw on knowledge across specialties and disease domains, including areas beyond their own expertise, to guide clinical decisions (**Figure 2a**).

**Clinical vignette (Table 1).** Consider a patient with a rare disease who presents with neurological and pulmonary symptoms. The patient is referred to a neurologist and a pulmonologist, each with deep expertise shaped by shared foundational training and system-level reasoning. Yet even broad training can be constrained by framing effects that narrow the differential diagnosis to more common or organ-specific causes. If the symptoms reflect a multisystem disorder, this framing may lead to missed connections or delayed diagnosis. Context-switching models can support clinicians by integrating knowledge across specialties, identifying shared pathophysiology, and generating a unified diagnosis and treatment plan. A related challenge arises in polypharmacy, when multiple physicians prescribe treatments for different conditions without visibility into overlapping risks or drug-drug interactions. Context-switching models can mitigate these issues by synthesizing cross-specialty knowledge, reasoning across comorbidities, and tailoring treatment to minimize therapeutic conflicts.

Context-switching across specialties and disease domains remains difficult because both care and model development often follow specialty-specific boundaries. Foundation models trained on siloed datasets or optimized for narrow objectives can reproduce the same limitations seen in human reasoning. Overcoming this requires models that perform multimodal reasoning and generalize across specialties, diseases, and combinations of data types, even when those combinations are absent during training.

*Data:* Without context-specific datasets, models cannot generalize across specialties and diseases. Domain expertise can be used to curate diverse instructions (for example, chain-of-thought reasoning steps that improve generalization to complex tasks) or response-preference sets (for example, human-annotated rankings of model outputs that align responses with user preferences) for each specialty or disease. A useful resource for enabling context switching is a prompt library: a structured collection of multimodal prompts linked to specific specialties and disease scenarios.

*Model:* Medical specialties rely on complementary data for diagnosis and treatment. For example, evaluating heart failure may require electrocardiograms, blood tests, and chest X-rays. In neurology, diagnosing a seizure disorder may involve electroencephalograms, brain imaging, and genetic testing. Multimodal models that identify and prioritize the most relevant data for a given clinical context, such as through context discovery (**Box 1**), can support context switching across specialties and diseases.

*Evaluation:* Model generalizability is critical for context switching. A model should not fail when faced with a new context that requires a novel combination of data modalities. This needs benchmarks that test context switching by assessing whether models can identify the correct contexts, generate responses relevant to the specialty or disease, and align outputs with specialist preferences[98–101] (for example, identifying context, generating relevant responses, and aligning outputs with specialist preferences). These benchmarks must also minimize data contamination (for example, overlap between pretraining and test datasets) to provide a valid measure of model generalizability[102,103].

## Tailoring care to geography and population

Understanding the incidence, prevalence, and mortality rates of health conditions in specific populations is essential for tailoring clinical decisions[104]. For example, isoniazid preventive therapy (IPT) is recommended by the World Health Organization for people living with HIV in high tuberculosis (TB) burden settings, even in the absence of active TB. However, there is ongoing debate about when to initiate IPT, especially at the time of HIV diagnosis, due to concerns about adherence, resistance, and regional variation in TB risk[105]. These decisions depend on global guidelines as well as on local epidemiology, healthcare infrastructure, and social context. For contextualized care, models need to interpret and apply guidelines in light of regional disease patterns, resource availability, and population needs[104,106,107] (**Figure 2b**).

**Clinical vignette (Table 1).** Consider the development of an international scoring function to assess organ or tissue function using patient data. A key risk is introducing contextual error by assigning different scores to patients with similar clinical profiles based only on geography or demographic attributes, such as age, sex, or race. These discrepancies often arise from uncalibrated models that ignore variation in healthcare delivery, disease prevalence, or access to care across settings. As a result, identical physiological inputs may yield different recommendations depending on context, leading to inappropriate care. Context-switching models address this by incorporating region-specific information, such as care practices, health infrastructure, or disease incidence, directly into score computation. A multi-center model can switch across country-level contexts to align with local systems and guidelines, while a multi-population model can detect subgroup-specific patterns and adapt outputs accordingly. For example, the model might recommend inpatient rather than outpatient treatment when follow-up care is difficult to access.

*Data:* Patient-level data vary substantially across settings, even within the same country. Differences in documentation, diagnostic workflows, and data completeness can limit the portability of models. Although regional systems and guidelines also influence care, the most immediate challenge is variability in the underlying data. To support context-switching, models must be trained on datasets that capture this heterogeneity. Engaging local experts and building regionally representative datasets enables models to capture both how care is delivered and why it varies across contexts.

*Model:* To generate context-specific recommendations, a reasoning model can integrate multi-step logic that captures both general clinical objectives and local constraints. This includes treatment-level decisions (for example, choosing a therapy based on symptoms) and fine-grained adjustments (for example, adapting dosing schedules or care settings to local resource availability). For instance, a model can be trained on human-annotated chain-of-thought reasoning traces to capture this nuanced multi-step decision-making process. A context-switching model can identify which factors vary across populations and incorporate them into its reasoning process to produce context-appropriate outputs.

*Evaluation:* Working with local and regional experts makes it possible to define metrics that evaluate whether model outputs align with guidelines and common practices in specific populations and geographies. Accuracy and robustness can also be measured by the trust of the target population in the model's recommendations and their acceptance of its use in the community[108].

## Democratizing access without reinforcing inequities

Access to medical care can be constrained by socioeconomic, geographic, financial, and cultural factors[109,110] (**Figure 2c**). Context switching can adapt to patient needs, care settings, and communication styles[111], but if models are built on biased data or without input from diverse communities, they risk reinforcing inequities. The impact of context switching depends on its design, who develops it, and whose context it reflects[112].

**Clinical vignette (Table 1).** Consider a patient who arrives at the emergency department with severe symptoms but frequently misses wellness visits. Although referred to oncology, the patient has not followed up. A typical response might be for the clinician to remind the patient to schedule the appointment, but this overlooks key contextual barriers. The patient may live far from a specialty clinic, lack reliable transportation or childcare, or be unable to take time off work. Without accounting for these constraints, the system risks recommending care that the patient cannot access: a contextual error.

Contextualized care incorporates these constraints. A more appropriate intervention can be to assign the patient a personalized medical chatbot for symptom monitoring and real-time feedback at home. The chatbot should communicate securely with the clinical team, augment documentation, and improve continuity of care. A context-switching conversational model can combine clinical knowledge with the communication styles and decision roles of different healthcare professionals to engage the patient effectively and collect relevant information. By adapting in this way, context-switching models can expand access to medical expertise and provide continuous, reliable points of contact between patients and their care teams.

*Data:* We can curate conversations between patients and healthcare professionals, including nurses, primary care physicians, specialists, as well as communications between clinicians. These conversations should reflect diverse populations, languages, and care settings to capture variation in both patient expression and expert communication styles. Additionally, human-annotated rankings of model outputs can ensure the alignment of generated responses with the healthcare role. Relevant clinical tools and databases can be integrated to improve the accuracy and adherence to clinical guidelines of the model.

*Model:* A context-switching conversational model can be structured as a generative agent system, where each agent embodies a distinct clinical persona. Agents can be selectively activated depending on the task or patient need, for example, a nurse practitioner for symptom triage or a specialist for treatment counseling. These agents retrieve and incorporate up-to-date patient records to adjust dialogue and reasoning in real-time based on evolving clinical context. Patients can initiate conversations with any agent or be directed to the appropriate one when uncertain. In collaboration with the intended users, guardrails and recovery policies can be implemented to maintain long-horizon consistency, reduce failure modes of subagents, and mitigate security and data governance violations.

*Evaluation:* Assessing whether these models democratize care requires metrics developed alongside medical and public health experts. One metric can be the rate of successful referrals to appropriate specialists, which reflects how well the model understands clinical roles. Appointment follow-through can serve as a proxy for patient engagement and perceived accessibility. Other relevant metrics can include symptom resolution timelines, quality of documentation, and patient-clinician satisfaction.

| **Clinical vignette** | **Contextual error**<br>Inappropriate care | **Contextualized care**<br>Appropriate care | **Context-switching model** |
|---|---|---|---|
| A patient has a rare disease with neurological and pulmonary symptoms. | The patient is referred to a neurologist and a pulmonologist. The neurologist and pulmonologist perform clinical evaluations that consider limited possibilities in a differential diagnosis. | A clinician leverages knowledge about all documented diseases to determine that the patient has a novel disease and identifies an underlying disease mechanism driving the presentation of symptoms in nervous and respiratory tissues. | **Medical specialty-switching model**<br><br>Augment each patient's medical records with existing biomedical knowledge, flexibly switching and combining disease contexts to arrive at a diagnosis and characterize the mechanism. |
| | | A clinician leverages knowledge about all documented diseases to identify a treatment plan for the patient that accounts for symptoms in both nervous and respiratory tissues while ensuring minimal polypharmacy effects. | **Disease-switching model**<br><br>Augment each patient's medical records with existing biomedical knowledge, flexibly switching and combining disease contexts to arrive at a multimorbidity-aware treatment plan that minimizes polypharmacy. |
| A clinician leads an international study on a new scoring metric computed using patients' medical data. | Patients with similar health profiles receive different scores due to uncalibrated models that do not account for regional disease prevalence or differences in care delivery. | The scoring metric is calibrated to reflect region-specific incidence rates and healthcare system constraints, leading to appropriate adjustments in predicted probabilities. | **Center-switching model**<br><br>Incorporate region-specific disease prevalence, healthcare practices, and resource availability to calibrate predictions appropriately. Switch between country or regional contexts to produce scores that reflect local conditions and standards of care. |
| | Patients receive different scores based on demographic attributes without clinical justification, reflecting bias in training data or modeling choices. | The scoring metric is calibrated using clinically relevant demographic factors, such as age, sex, or genetic ancestry, when they are medically appropriate to include. | **Population-switching model**<br><br>Automatically recognize clinically relevant demographic patterns, such as age-related risks or ancestry-linked biomarkers, and adjust predictions based on population-specific health factors while avoiding unjustified bias or overfitting. |
| A patient with severe symptoms goes to the ER. The patient frequently misses wellness visits and has not seen a specialist despite an oncology referral. | The clinician reminds the patient to make an appointment with the oncologist. | The clinician shows the patient a personalized medical chatbot to monitor symptoms at home with real-time feedback using the conversational medical model. The conversations are recorded and shared confidentially and securely with the patient's medical team (including the referred oncologist) to inform future visits. | **Healthcare role-switching model**<br><br>Learn the linguistic patterns of diverse clinical personas alongside comprehensive medical knowledge to engage with patients, collect information about their health, and triage patients. |

**Table 1. Clinical vignettes of context-switching in medical AI.** Each vignette illustrates how a lack of contextual adaptation can lead to inappropriate care. We envision context-switching models that adjust across specialties, diseases, geographic settings, patient populations, and clinical roles.

## Conclusion

We envision context-switching as the defining paradigm of medical AI. Rather than building separate models for each specialty or health system, future models must adjust their reasoning at test time to the data, clinicians, patients, and decisions required in context. This shift moves medical AI beyond narrow, static tools toward models that reflect the complexity and variability of clinical practice. Medical care spans diverse specialties, populations, and workflows that differ across settings and evolve over time. To remain reliable, models must respond to this variation without manual adjustment. Context-switching models should adapt in real-time, reduce dependence on curated data, and support consistent performance across health systems. Progress must move beyond benchmarking to advance reasoning, generalization, and adaptation to clinical goals. Only context-switching models can scale to match the complexity of practice and the diversity of patients.

## Competing interests

The authors declare no competing interests.

## Acknowledgements

M.M.L. is supported by The Ivan and Francesca Berkowitz Family Living Laboratory Collaboration at Harvard Medical School and Clalit Research Institute. We gratefully acknowledge the support of NIH R01-HD108794, NSF CAREER 2339524, US DoD FA8702-15-D-0001, ARPA-H BDF program, awards from Chan Zuckerberg Initiative, Bill & Melinda Gates Foundation INV-079038, Amazon Faculty Research, Google Research Scholar Program, AstraZeneca Research, Roche Alliance with Distinguished Scientists, Sanofi iDEA-iTECH, Pfizer Research, John and Virginia Kaneb Fellowship at Harvard Medical School, Biswas Computational Biology Initiative in partnership with the Milken Institute, Harvard Medical School Dean's Innovation Fund for the Use of Artificial Intelligence, Harvard Data Science Initiative, and Kempner Institute for the Study of Natural and Artificial Intelligence at Harvard University. Any opinions, findings, conclusions or recommendations expressed in this material are those of the authors and do not necessarily reflect the views of the funders.